\documentclass[runningheads]{llncs}
\usepackage{graphicx}
\usepackage{amsmath,amssymb} 
\usepackage{color}

\usepackage{booktabs}
\usepackage[table]{xcolor}
\usepackage{multirow}

\usepackage[breaklinks=true,colorlinks,bookmarks=false]{hyperref}

\newcommand{\bm}[1]{\boldsymbol{#1}}

\makeatletter
\usepackage{xspace}
\def\@onedot{\ifx\@let@token.\else.\null\fi\xspace}
\DeclareRobustCommand\onedot{\futurelet\@let@token\@onedot}

\newcommand{\figref}[1]{Fig\onedot~\ref{#1}}

\newcommand{\secref}[1]{Sec\onedot~\ref{#1}}
\newcommand{\tabref}[1]{Tab\onedot~\ref{#1}}

\def\eg{\emph{e.g}\onedot} 
\def\ie{\emph{i.e}\onedot} 
 
 \def\vs{\emph{vs}\onedot}

\begin{document}
\title{Encoder-Decoder with Atrous Separable Convolution for Semantic Image Segmentation} 

\titlerunning{DeepLabv3+: Encoder-Decoder with Atrous Separable Convolution}
%

\author{Liang-Chieh~Chen \and Yukun~Zhu \and George~Papandreou \and Florian~Schroff \and Hartwig~Adam}

%
\authorrunning{L.-C Chen, Y. Zhu, G. Papandreou, F. Schroff, and H. Adam}
%

\institute{Google Inc.\\
\email{\{lcchen, yukun, gpapan, fschroff, hadam\}@google.com}}
\maketitle              

\begin{abstract}
  Spatial pyramid pooling module or encode-decoder structure are used in deep neural networks for semantic segmentation task. The former networks are able to encode multi-scale contextual information by probing the incoming features with filters or pooling operations at multiple rates and multiple effective fields-of-view, while the latter networks can capture sharper object boundaries by gradually recovering the spatial information. In this work, we propose to combine the advantages from both methods. Specifically, our proposed model, DeepLabv3+, extends DeepLabv3 by adding a simple yet effective decoder module to refine the segmentation results especially along object boundaries. We further explore the Xception model and apply the depthwise separable convolution to both Atrous Spatial Pyramid Pooling and decoder modules, resulting in a faster and stronger encoder-decoder network. We demonstrate the effectiveness of the proposed model on PASCAL VOC 2012 and Cityscapes datasets, achieving the test set performance of 89.0\% and 82.1\% without any post-processing. Our paper is accompanied with a publicly available reference implementation of the proposed models in Tensorflow at \url{https://github.com/tensorflow/models/tree/master/research/deeplab}.
\keywords{Semantic image segmentation, spatial pyramid pooling, encoder-decoder, and depthwise separable convolution.}
\end{abstract}

\section{Introduction}

\begin{figure}[!t]
  \centering
  \begin{tabular}{c c c}
    \includegraphics[height=0.33\linewidth]{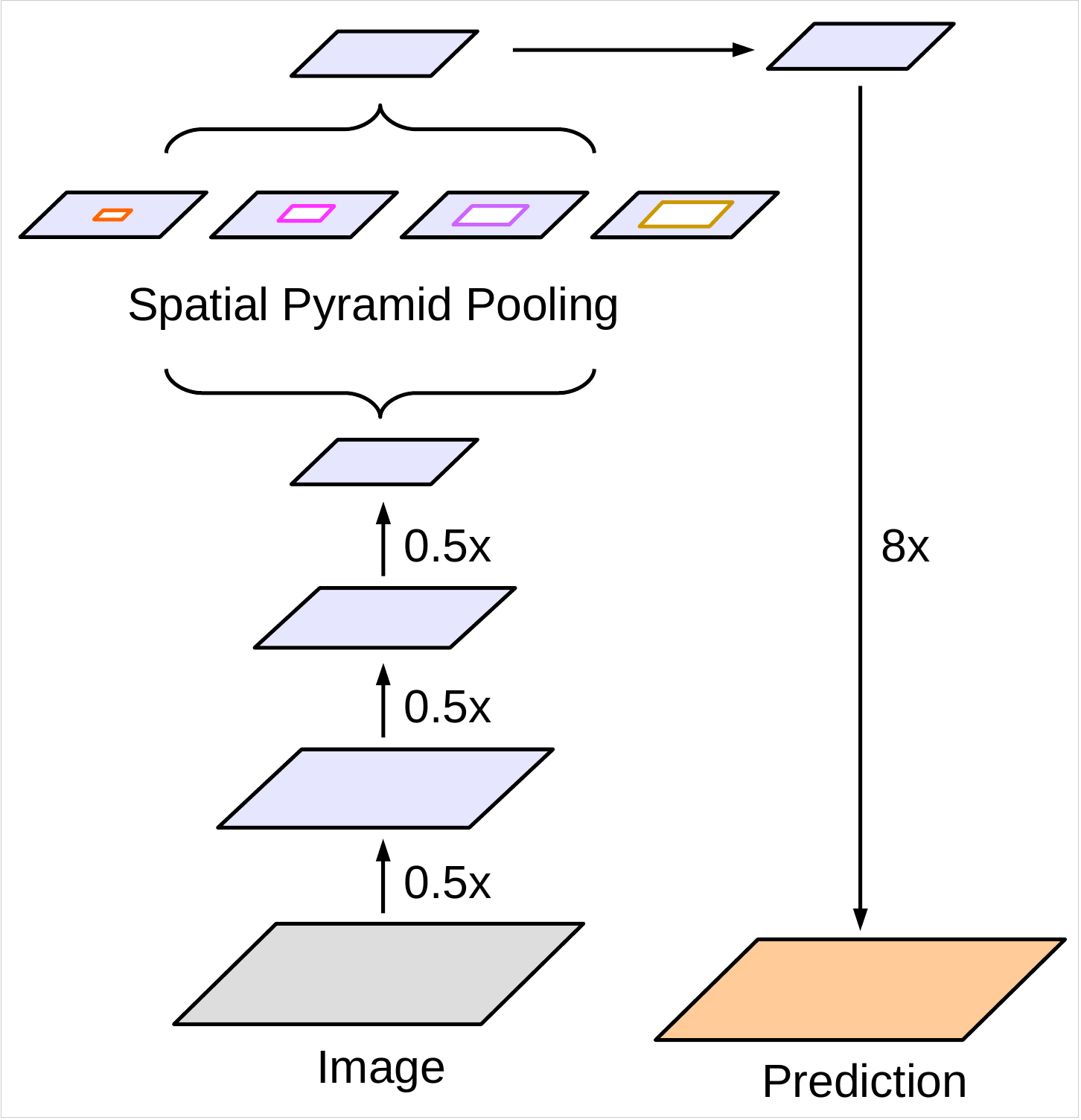} &
    \includegraphics[height=0.33\linewidth]{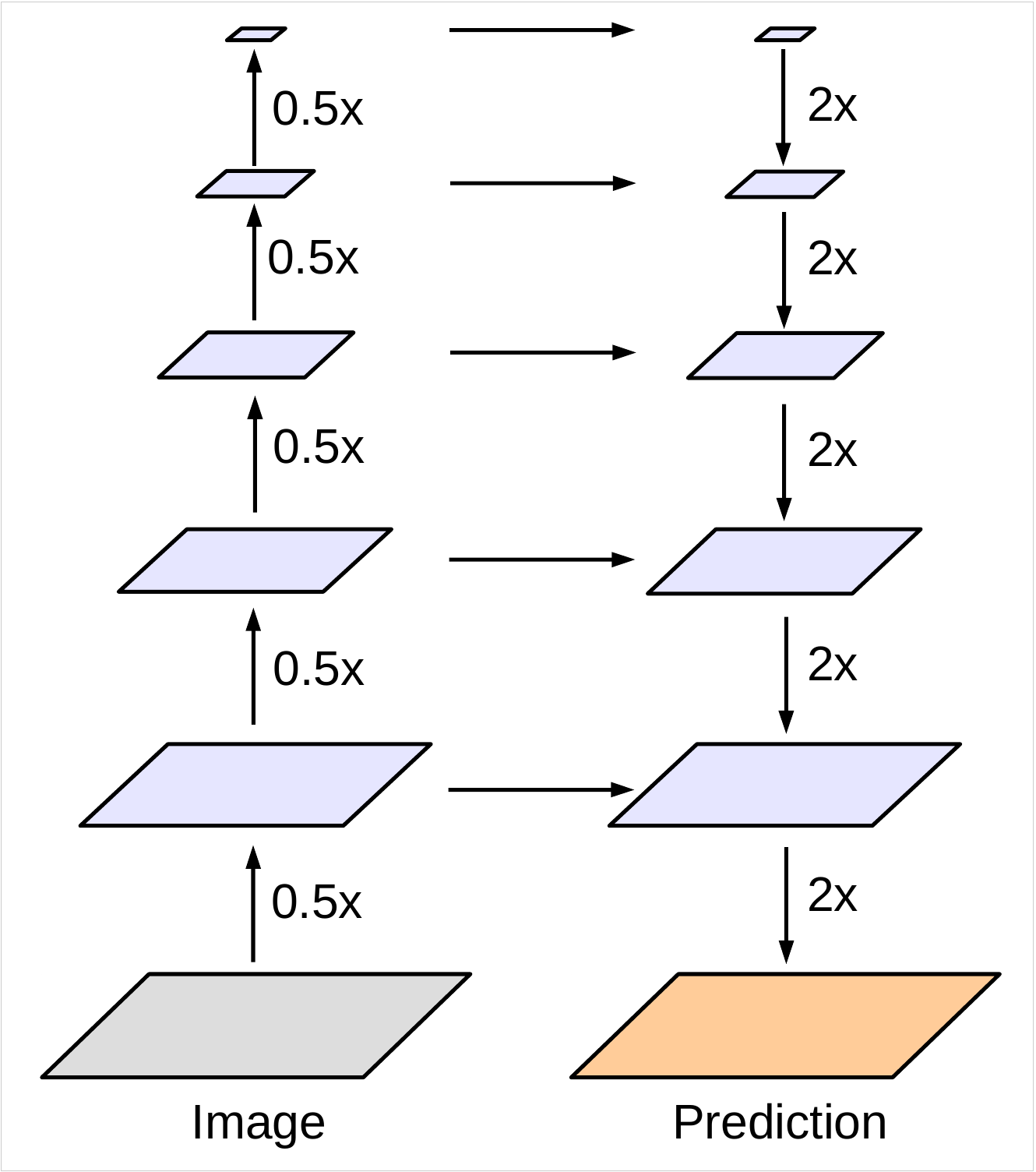} &
    \includegraphics[height=0.33\linewidth]{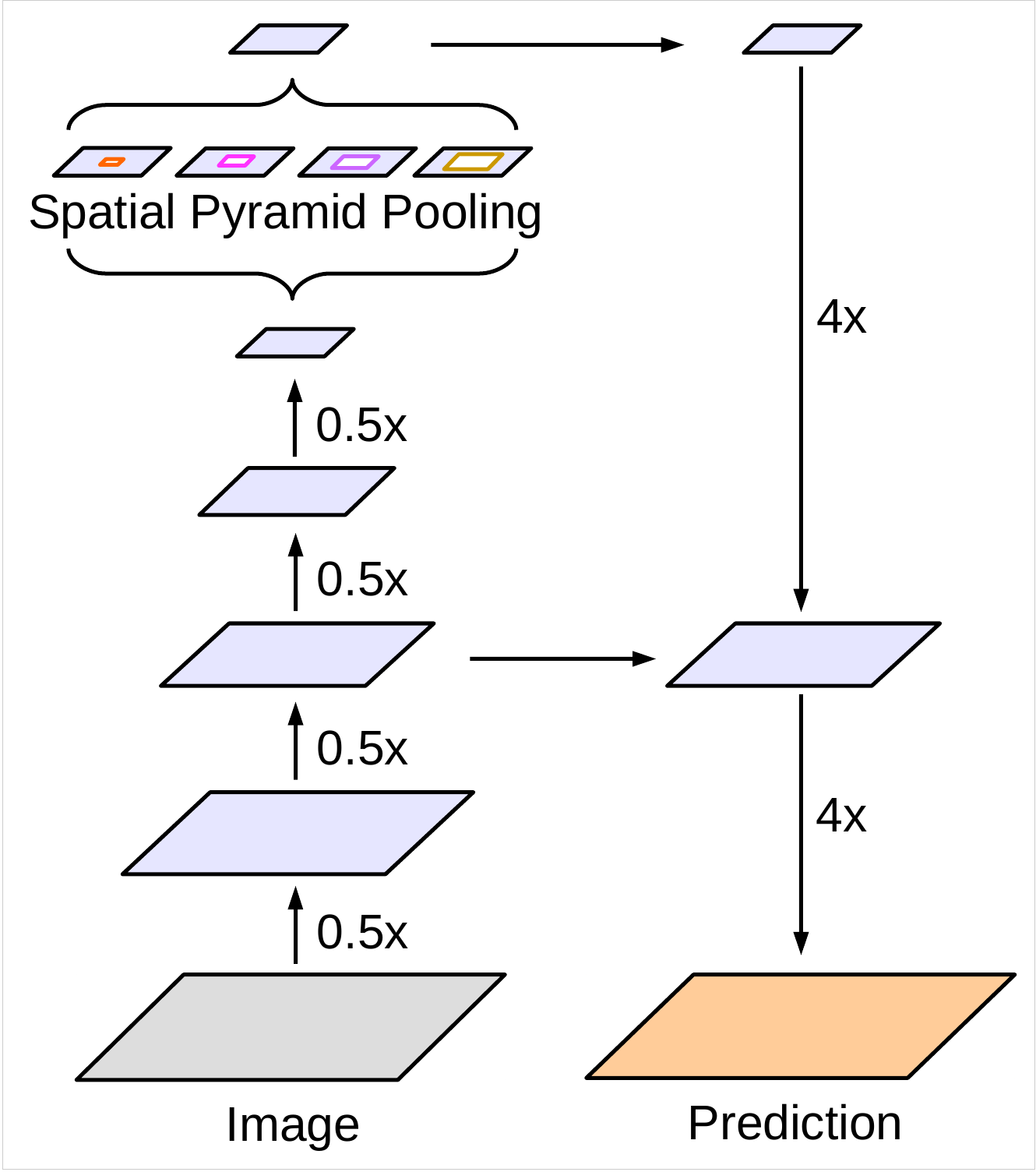} \\
    {\tiny (a) Spatial Pyramid Pooling} &
    {\tiny (b) Encoder-Decoder} &
    {\tiny (c) Encoder-Decoder with Atrous Conv} \\
  \end{tabular}
  \caption{We improve DeepLabv3, which employs the spatial pyramid pooling module (a), with the encoder-decoder structure (b). The proposed model, DeepLabv3+, contains rich semantic information from the encoder module, while the detailed object boundaries are recovered by the simple yet effective decoder module. The encoder module allows us to extract features at an arbitrary resolution by applying atrous convolution.}
  \label{fig:overiew}
\end{figure}

Semantic segmentation with the goal to assign semantic labels to every pixel in an image \cite{everingham2014pascal,mottaghi2014role,Cordts2016Cityscapes,zhou2017scene,caesar2016cocostuff} is one of the fundamental topics in computer vision. Deep convolutional neural networks \cite{LeCun1998,krizhevsky2012imagenet,sermanet2013overfeat,simonyan2014very,szegedy2014going} based on the Fully Convolutional Neural Network \cite{sermanet2013overfeat,long2014fully} show striking improvement over systems relying on hand-crafted features \cite{he2004multiscale,shotton2009textonboost,kohli2009robust,ladicky2009associative,gould2009decomposing,yao2012describing} on benchmark tasks. In this work, we consider two types of neural networks that use spatial pyramid pooling module \cite{grauman2005pyramid,lazebnik2006beyond,he2014spatial} or encoder-decoder structure \cite{ronneberger2015u,badrinarayanan2015segnet} for semantic segmentation, where the former one captures rich contextual information by pooling features at different resolution while the latter one is able to obtain sharp object boundaries.

In order to capture the contextual information at multiple scales, DeepLabv3 \cite{chen2017rethinking} applies several parallel atrous convolution with different rates (called Atrous Spatial Pyramid Pooling, or ASPP), while PSPNet \cite{zhao2017pyramid} performs pooling operations at different grid scales. Even though rich semantic information is encoded in the last feature map, detailed information related to object boundaries is missing due to the pooling or convolutions with striding operations within the network backbone. This could be alleviated by applying the atrous convolution to extract denser feature maps. However, given the design of state-of-art neural networks \cite{krizhevsky2012imagenet,simonyan2014very,szegedy2014going,he2015deep,chollet2016xception} and limited GPU memory, it is computationally prohibitive to extract output feature maps that are 8, or even 4 times smaller than the input resolution. Taking ResNet-101 \cite{he2015deep} for example, when applying atrous convolution to extract output features that are 16 times smaller than input resolution, features within the last 3 residual blocks (9 layers) have to be dilated. Even worse, \textbf{26} residual blocks (\textbf{78} layers!) will be affected if output features that are 8 times smaller than input are desired. Thus, it is computationally intensive if denser output features are extracted for this type of models. On the other hand, encoder-decoder models \cite{ronneberger2015u,badrinarayanan2015segnet} lend themselves to faster computation (since no features are dilated) in the encoder path and gradually recover sharp object boundaries in the decoder path. Attempting to combine the advantages from both methods, we propose to enrich the encoder module in the encoder-decoder networks by incorporating the multi-scale contextual information.

In particular, our proposed model, called DeepLabv3+, extends DeepLabv3 \cite{chen2017rethinking} by adding a simple yet effective decoder module to recover the object boundaries, as illustrated in \figref{fig:overiew}. The rich semantic information is encoded in the output of DeepLabv3, with atrous convolution allowing one to control the density of the encoder features, depending on the budget of computation resources. Furthermore, the decoder module allows detailed object boundary recovery.

Motivated by the recent success of depthwise separable convolution \cite{sifre2014rigid,vanhoucke2014learning,chollet2016xception,howard2017mobilenets,zhang2017shufflenets}, we also explore this operation and show improvement in terms of both speed and accuracy by adapting the Xception model \cite{chollet2016xception}, similar to \cite{dai2017coco}, for the task of semantic segmentation, and applying the atrous separable convolution to both the ASPP and decoder modules. Finally, we demonstrate the effectiveness of the proposed model on PASCAL VOC 2012 and Cityscapes datasts and attain the test set performance of 89.0\% and 82.1\% without any post-processing, setting a new state-of-the-art.

In summary, our contributions are:
\begin{itemize}
  \item We propose a novel encoder-decoder structure which employs DeepLabv3 as a powerful encoder module and a simple yet effective decoder module.
  \item In our structure, one can arbitrarily control the resolution of extracted encoder features by atrous convolution to trade-off precision and runtime, which is not possible with existing encoder-decoder models.
  \item We adapt the Xception model for the segmentation task and apply depthwise separable convolution to both ASPP module and decoder module, resulting in a faster and stronger encoder-decoder network.
  \item Our proposed model attains a new state-of-art performance on PASCAL VOC 2012 and Cityscapes datasets. We also provide detailed analysis of design choices and model variants.
  \item We make our Tensorflow-based implementation of the proposed model publicly available at \url{https://github.com/tensorflow/models/tree/master/research/deeplab}.
\end{itemize}

\section{Related Work}
Models based on Fully Convolutional Networks (FCNs) \cite{sermanet2013overfeat,long2014fully} have demonstrated significant improvement on several segmentation benchmarks \cite{everingham2014pascal,mottaghi2014role,Cordts2016Cityscapes,zhou2017scene,caesar2016cocostuff}. There are several model variants proposed to exploit the contextual information for segmentation \cite{he2004multiscale,shotton2009textonboost,kohli2009robust,ladicky2009associative,gould2009decomposing,yao2012describing,mostajabi2014feedforward,dai2015convolutional}, including those that employ multi-scale inputs (\ie, image pyramid) \cite{farabet2013learning,eigen2015predicting,pinheiro2014recurrent,lin2015efficient,chen2015attention,chen2017deeplab} or those that adopt probabilistic graphical models (such as DenseCRF \cite{krahenbuhl2011efficient} with efficient inference algorithm \cite{adams2010fast}) \cite{chen2014semantic,bell2014material,zheng2015conditional,lin2015efficient,liu2015semantic,papandreou2015weakly,schwing2015fully,jampani2016learning,Vemulapalli2016Gaussian,chandra2016fast,chandra2017dense,chen2017deeplab}. In this work, we mainly discuss about the models that use spatial pyramid pooling and encoder-decoder structure.

\textbf{Spatial pyramid pooling:} Models, such as PSPNet \cite{zhao2017pyramid} or DeepLab \cite{chen2017deeplab,chen2017rethinking}, perform spatial pyramid pooling \cite{grauman2005pyramid,lazebnik2006beyond} at several grid scales (including image-level pooling \cite{liu2015parsenet}) or apply several parallel atrous convolution with different rates (called Atrous Spatial Pyramid Pooling, or ASPP). These models have shown promising results on several segmentation benchmarks by exploiting the multi-scale information.

\textbf{Encoder-decoder:} The encoder-decoder networks have been successfully applied to many computer vision tasks, including human pose estimation \cite{newell2016stacked}, object detection \cite{lin2016feature,shrivastava2016beyond,fu2017dssd}, and semantic segmentation \cite{long2014fully,noh2015learning,ronneberger2015u,badrinarayanan2015segnet,lin2016refinenet,pohlen2016full,peng2017large,islamgated,wojna2017devil,fu2017stacked,zhang2018exfuse}. Typically, the encoder-decoder networks contain (1) an encoder module that gradually reduces the feature maps and captures higher semantic information, and (2) a decoder module that gradually recovers the spatial information. Building on top of this idea, we propose to use DeepLabv3 \cite{chen2017rethinking} as the encoder module and add a simple yet effective decoder module to obtain sharper segmentations.

\textbf{Depthwise separable convolution:} Depthwise separable convolution \cite{sifre2014rigid,vanhoucke2014learning} or group convolution \cite{krizhevsky2012imagenet,xie2017aggreated}, a powerful operation to reduce the computation cost and number of parameters while maintaining similar (or slightly better) performance. This operation has been adopted in many recent neural network designs \cite{jin2014flattened,wang2016factorized,chollet2016xception,howard2017mobilenets,zhang2017shufflenets,dai2017coco,zoph2017learning}. In particular, we explore the Xception model \cite{chollet2016xception}, similar to \cite{dai2017coco} for their COCO 2017 detection challenge submission, and show improvement in terms of both accuracy and speed for the task of semantic segmentation.

\section{Methods}
\label{sec:methods}
In this section, we briefly introduce atrous convolution \cite{holschneider1989real,giusti2013fast,sermanet2013overfeat,papandreou2014untangling,chen2014semantic} and depthwise separable convolution \cite{sifre2014rigid,vanhoucke2014learning,wang2016factorized,chollet2016xception,howard2017mobilenets}. We then review DeepLabv3 \cite{chen2017rethinking} which is used as our encoder module before discussing the proposed decoder module appended to the encoder output. We also present a modified Xception model \cite{chollet2016xception,dai2017coco} which further improves the performance with faster computation.

\subsection{Encoder-Decoder with Atrous Convolution}
\label{subsec:encoder_decoder}

\begin{figure}[!t]
  \centering
  \begin{tabular}{c}
    \includegraphics[width=0.96\linewidth]{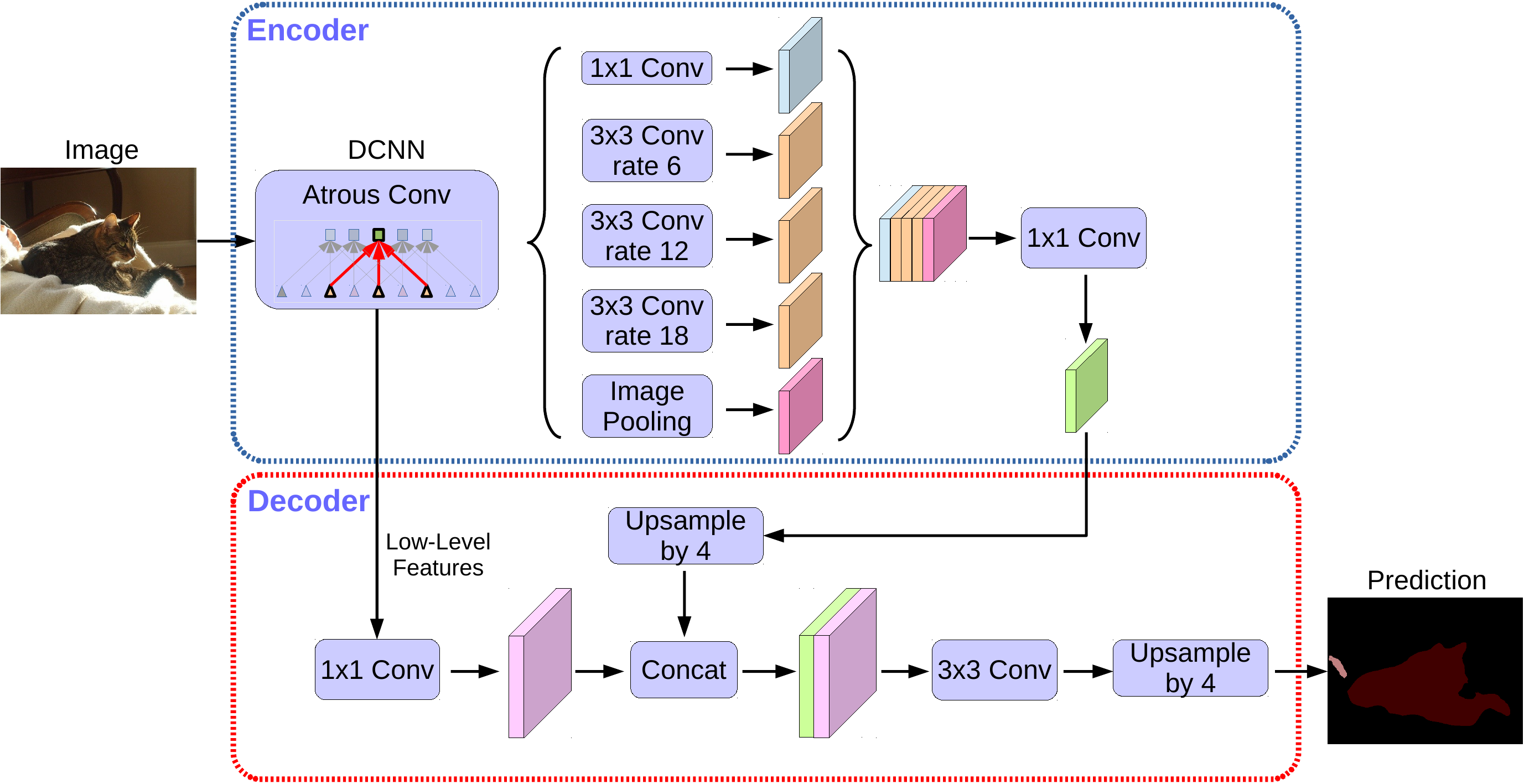} \\
  \end{tabular}
  \caption{Our proposed DeepLabv3+ extends DeepLabv3 by employing a encoder-decoder structure. The encoder module encodes multi-scale contextual information by applying atrous convolution at multiple scales, while the simple yet effective decoder module refines the segmentation results along object boundaries.}
  \label{fig:model}
\end{figure}

{\bf Atrous convolution:} Atrous convolution, a powerful tool that allows us to explicitly control the resolution of features computed by deep convolutional neural networks and adjust filter's field-of-view in order to capture multi-scale information, generalizes standard convolution operation. In the case of two-dimensional signals, for each location $\bm{i}$ on the output feature map $\bm{y}$ and a convolution filter $\bm{w}$, atrous convolution is applied over the input feature map $\bm{x}$ as follows:

\begin{equation}
  \bm{y}[\bm{i}] = \sum_{\bm{k}} \bm{x}[\bm{i} + r \cdot \bm{k}] \bm{w}[\bm{k}]
\end{equation}
where the atrous rate \emph{r} determines the stride with which we sample the input signal. We refer interested readers to \cite{chen2017deeplab} for more details. Note that standard convolution is a special case in which rate $r = 1$. The filter's field-of-view is adaptively modified by changing the rate value.

\begin{figure}[!t]
  \centering
  \scalebox{1}{
  \begin{tabular}{c c c}
    \includegraphics[width=0.3\linewidth]{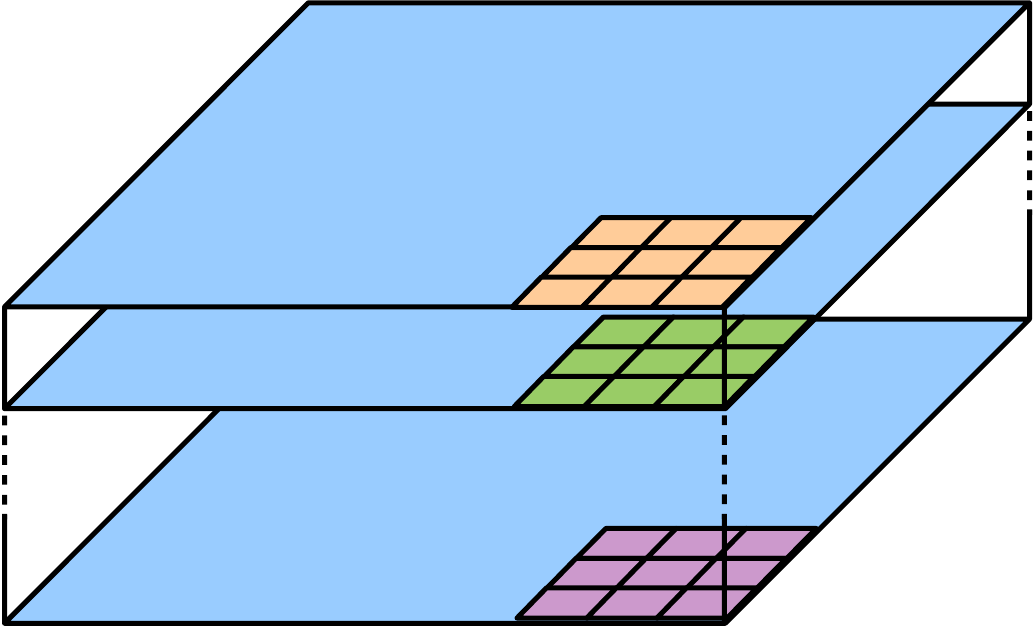} &
    \includegraphics[width=0.3\linewidth]{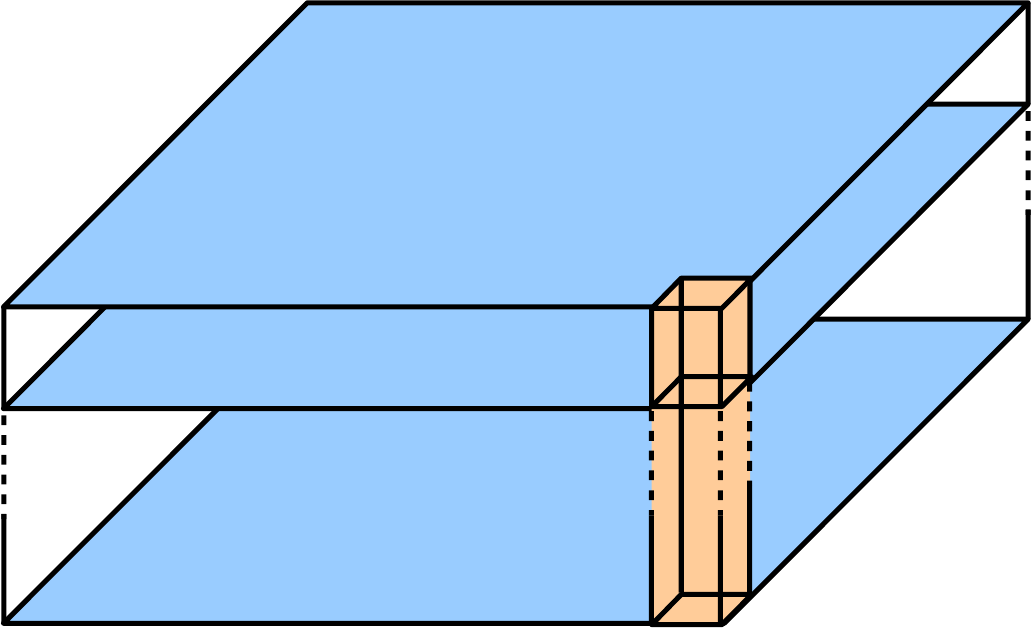} &
    \includegraphics[width=0.3\linewidth]{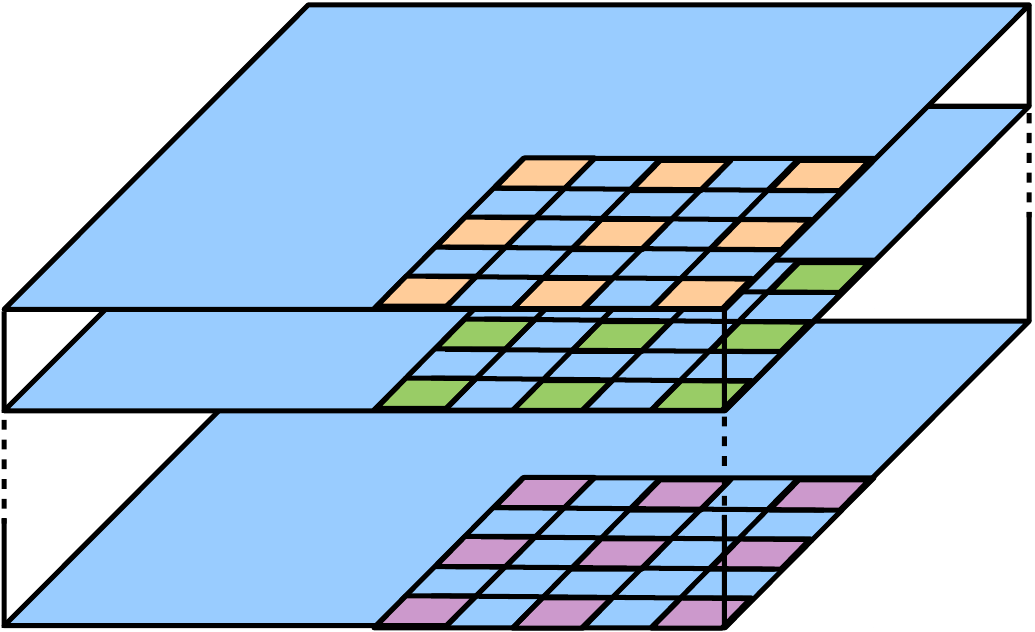} \\
    (a) Depthwise conv. &
    (b) Pointwise conv. &
    (c) Atrous depthwise conv. \\
  \end{tabular}
  }
  \caption{$3\times3$ Depthwise separable convolution decomposes a standard convolution into (a) a depthwise convolution (applying a single filter for each input channel) and (b) a pointwise convolution (combining the outputs from depthwise convolution across channels). In this work, we explore \textit{atrous separable convolution} where atrous convolution is adopted in the depthwise convolution, as shown in (c) with $rate=2$.}
  \label{fig:atrous_separable_conv}
\end{figure}

{\bf Depthwise separable convolution:} Depthwise separable convolution, factorizing a standard convolution into a \textit{depthwise convolution} followed by a \textit{pointwise convolution} (\ie, $1\times1$ convolution), drastically reduces computation complexity. Specifically, the depthwise convolution performs a spatial convolution independently for each input channel, while the pointwise convolution is employed to combine the output from the depthwise convolution. In the TensorFlow \cite{abadi2016tensorflow} implementation of depthwise separable convolution, atrous convolution has been supported in the depthwise convolution (\ie, the spatial convolution), as illustrated in \figref{fig:atrous_separable_conv}. In this work, we refer the resulting convolution as \textit{atrous separable convolution}, and found that atrous separable convolution significantly reduces the computation complexity of proposed model while maintaining similar (or better) performance.

{\bf DeepLabv3 as encoder:} DeepLabv3 \cite{chen2017rethinking} employs atrous convolution \cite{holschneider1989real,giusti2013fast,sermanet2013overfeat,papandreou2014untangling} to extract the features computed by deep convolutional neural networks at an arbitrary resolution. Here, we denote $\emph{output\ stride}$ as the ratio of input image spatial resolution to the final output resolution (before global pooling or fully-connected layer). For the task of image classification, the spatial resolution of the final feature maps is usually 32 times smaller than the input image resolution and thus $\emph{output\ stride}=32$. For the task of semantic segmentation, one can adopt $\emph{output\ stride}=16$ (or $8$) for denser feature extraction by removing the striding in the last one (or two) block(s) and applying the atrous convolution correspondingly (\eg, we apply $rate=2$ and $rate=4$ to the last two blocks respectively for $\emph{output\ stride}=8$). Additionally, DeepLabv3 augments the Atrous Spatial
Pyramid Pooling module, which probes convolutional features at multiple scales by applying atrous convolution with different rates, with the image-level features \cite{liu2015parsenet}. We use the last feature map before logits in the original DeepLabv3 as the encoder output in our proposed encoder-decoder structure. Note the encoder output feature map contains 256 channels and rich semantic information. Besides, one could extract features at an arbitrary resolution by applying the atrous convolution, depending on the computation budget.

{\bf Proposed decoder:} The encoder features from DeepLabv3 are usually computed with $\emph{output\ stride}=16$. In the work of \cite{chen2017rethinking}, the features are bilinearly upsampled by a factor of 16, which could be considered a naive decoder module. However, this naive decoder module may not successfully recover object segmentation details. We thus propose a simple yet effective decoder module, as illustrated in \figref{fig:model}. The encoder features are first bilinearly upsampled by a factor of 4 and then concatenated with the corresponding low-level features \cite{hariharan2014hypercolumns} from the network backbone that have the same spatial resolution (\eg, Conv2 before striding in ResNet-101 \cite{he2015deep}). We apply another $1\times1$ convolution on the low-level features to reduce the number of channels, since the corresponding low-level features usually contain a large number of channels (\eg, 256 or 512) which may outweigh the importance of the rich encoder features (only 256 channels in our model) and make the training harder. After the concatenation, we apply a few $3\times3$ convolutions to refine the features followed by another simple bilinear upsampling by a factor of 4. We show in \secref{sec:experiments} that using $\emph{output\ stride}=16$ for the encoder module strikes the best trade-off between speed and accuracy. The performance is marginally improved when using $\emph{output\ stride}=8$ for the encoder module at the cost of extra computation complexity.

\begin{figure}[!t]
  \centering
  \scalebox{1}{
  \begin{tabular}{c c}
    \includegraphics[width=0.46\linewidth]{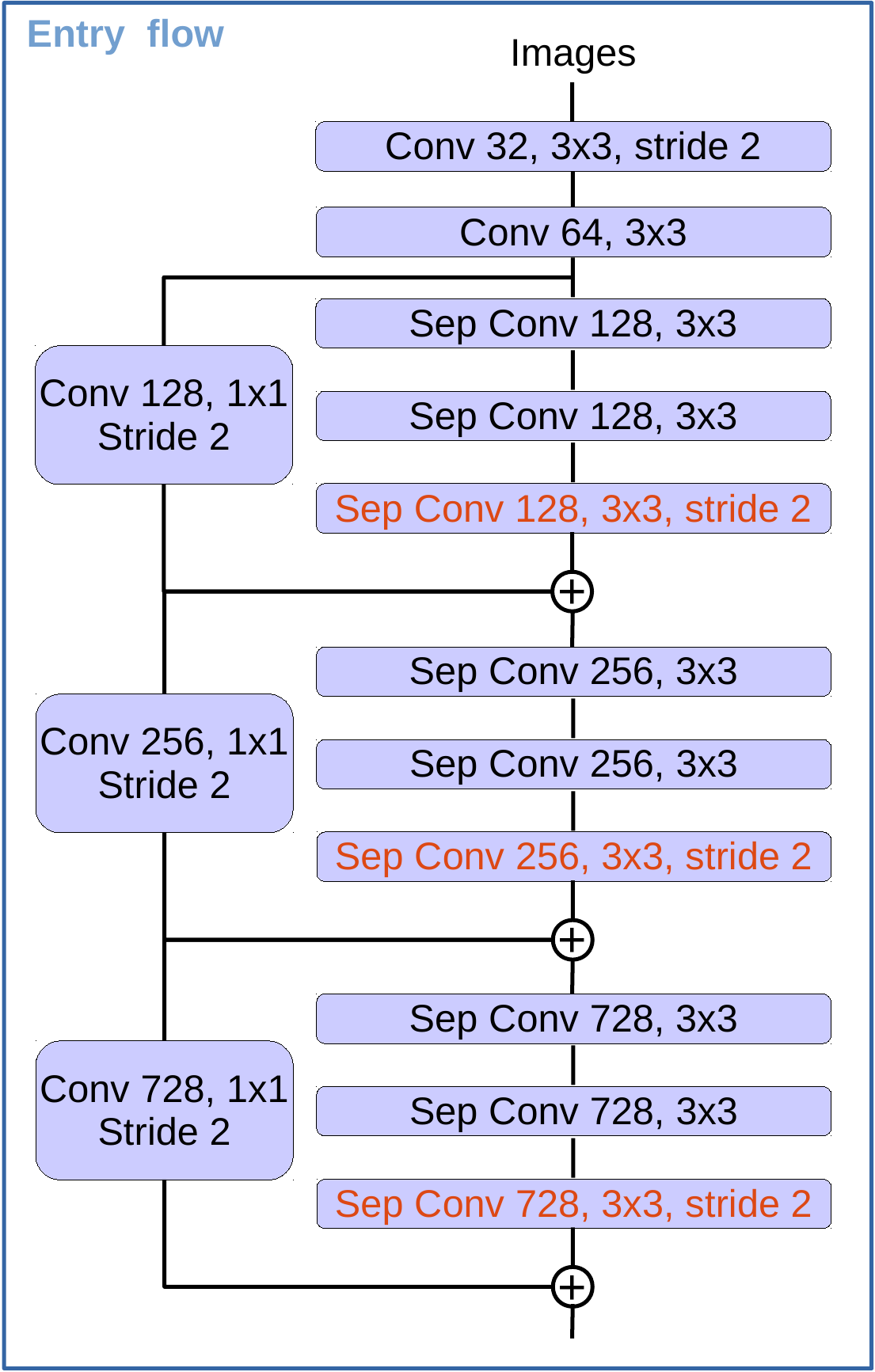} &
    \includegraphics[width=0.46\linewidth]{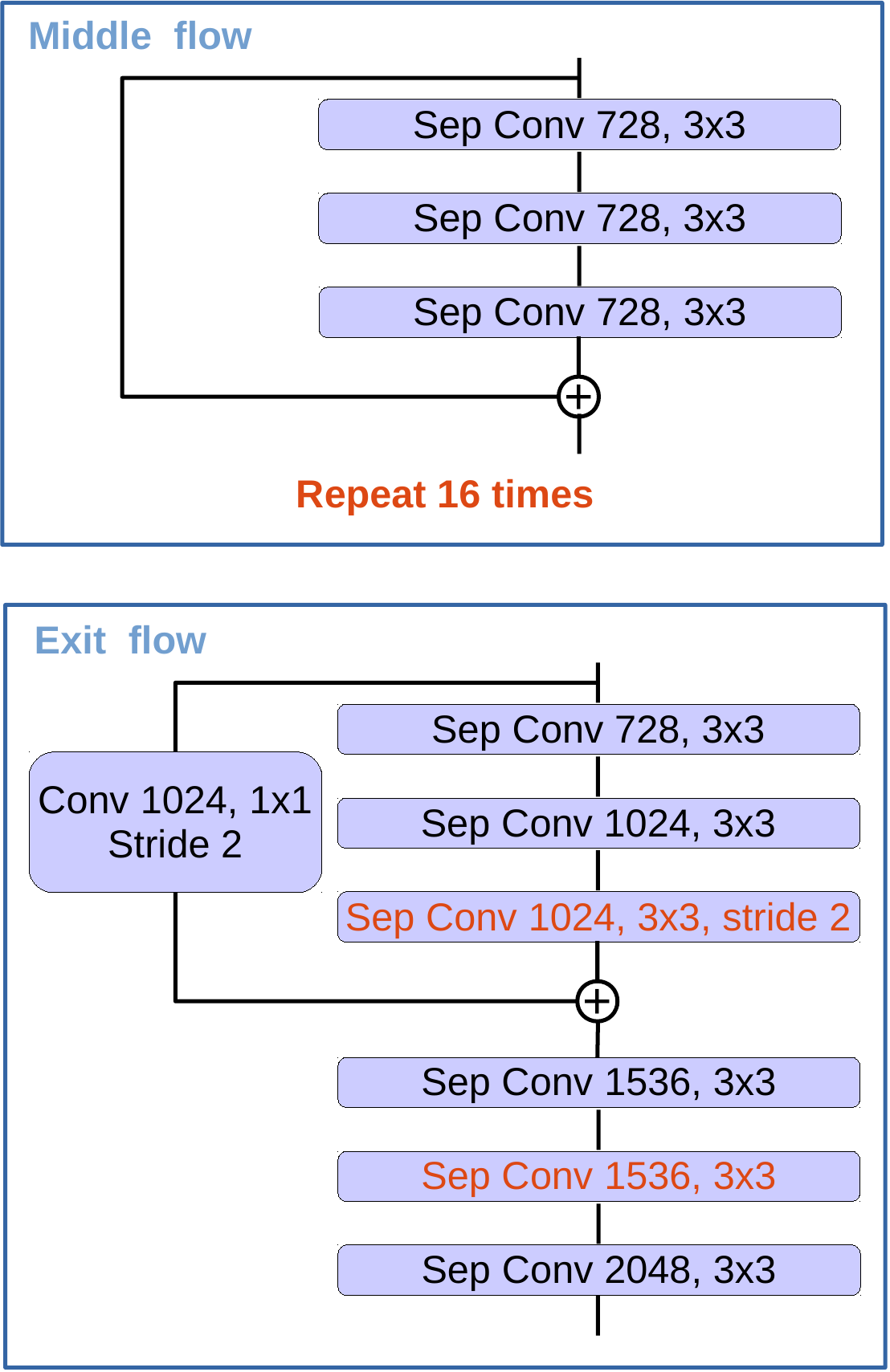} \\
  \end{tabular}
  }
  \caption{We modify the Xception as follows: (1) more layers (same as MSRA's modification except the changes in Entry flow), (2) all the max pooling operations are replaced by depthwise separable convolutions with striding, and (3) extra batch normalization and ReLU are added after each $3\times3$ depthwise convolution, similar to MobileNet.}
  \label{fig:xception}
\end{figure}

\subsection{Modified Aligned Xception}
\label{subsec:xception}

The Xception model \cite{chollet2016xception} has shown promising image classification results on ImageNet \cite{ILSVRC15} with fast computation. More recently, the MSRA team \cite{dai2017coco} modifies the Xception model (called Aligned Xception) and further pushes the performance in the task of object detection. Motivated by these findings, we work in the same direction to adapt the Xception model for the task of semantic image segmentation. In particular, we make a few more changes on top of MSRA's modifications, namely (1) deeper Xception same as in \cite{dai2017coco} except that we do not modify the entry flow network structure for fast computation and memory efficiency, (2) all max pooling operations are replaced by depthwise separable convolution with striding, which enables us to apply \textit{atrous separable convolution} to extract feature maps at an arbitrary resolution (another option is to extend the atrous algorithm to max pooling operations), and (3) extra batch normalization \cite{ioffe2015batch} and ReLU activation are added after each $3\times3$ depthwise convolution, similar to MobileNet design \cite{howard2017mobilenets}. See \figref{fig:xception} for details.

\section{Experimental Evaluation}
\label{sec:experiments}

We employ ImageNet-1k \cite{ILSVRC15} pretrained ResNet-101 \cite{he2015deep} or modified aligned Xception \cite{chollet2016xception,dai2017coco} to extract dense feature maps by atrous convolution. Our implementation is built on TensorFlow \cite{abadi2016tensorflow} and is made publicly available.


The proposed models are evaluated on the PASCAL VOC 2012 semantic segmentation benchmark \cite{everingham2014pascal} which contains 20 foreground object classes and one background class. The original dataset contains $1,464$ (\textit{train}), $1,449$ (\textit{val}), and $1,456$ (\textit{test}) pixel-level annotated
images. We augment the dataset
by the extra annotations provided by \cite{hariharan2011semantic},
resulting in $10,582$ (\textit{trainaug}) training images. The performance
is measured in terms of pixel intersection-over-union averaged across
the 21 classes (mIOU).

We follow the same training protocol as in \cite{chen2017rethinking} and refer the interested readers to \cite{chen2017rethinking} for details. In short, we employ the same learning rate schedule (\ie, ``poly'' policy \cite{liu2015parsenet} and same initial learning rate $0.007$), crop size $513\times513$, fine-tuning batch normalization parameters \cite{ioffe2015batch} when $\emph{output\ stride}=16$, and random scale data augmentation during training. Note that we also include batch normalization parameters in the proposed decoder module. Our proposed model is trained end-to-end without piecewise pretraining of each component.

\subsection{Decoder Design Choices}
We define ``DeepLabv3 feature map'' as the last feature map computed by DeepLabv3 (\ie, the features containing ASPP features and image-level features), and $[k\times k, f]$ as a convolution operation with kernel $k\times k$ and $f$ filters.

When employing $\emph{output\ stride}=16$, ResNet-101 based DeepLabv3 \cite{chen2017rethinking} bilinearly upsamples the logits by $16$ during both training and evaluation. This simple bilinear upsampling could be considered as a naive decoder design, attaining the performance of $77.21\%$ \cite{chen2017rethinking} on PASCAL VOC 2012 \textit{val} set and is $1.2\%$ better than not using this naive decoder during training (\ie, downsampling groundtruth during training). To improve over this naive baseline, our proposed model ``DeepLabv3+'' adds the decoder module on top of the encoder output, as shown in \figref{fig:model}. In the decoder module, we consider three places for different design choices, namely (1) the $1\times1$ convolution used to reduce the channels of the low-level feature map from the encoder module, (2) the $3\times3$ convolution used to obtain sharper segmentation results, and (3) what encoder low-level features should be used.

To evaluate the effect of the $1\times1$ convolution in the decoder module, we employ $[3\times3, 256]$ and the Conv2 features from ResNet-101 network backbone, \ie, the last feature map in res2x residual block (to be concrete, we use the feature map before striding). As shown in \tabref{tab:decoder_bottleneck}, reducing the channels of the low-level feature map from the encoder module to either 48 or 32 leads to better performance. We thus adopt $[1\times1, 48]$ for channel reduction.

We then design the $3\times3$ convolution structure for the decoder module and report the findings in \tabref{tab:decoder_structure}. We find that after concatenating the Conv2 feature map (before striding) with DeepLabv3 feature map, it is more effective to employ two $3\times3$ convolution with 256 filters than using simply one or three convolutions. Changing the number of filters from 256 to 128 or the kernel size from $3\times3$ to $1\times1$ degrades performance. We also experiment with the case where both Conv2 and Conv3 feature maps are exploited in the decoder module. In this case, the decoder feature map are gradually upsampled by 2, concatenated with Conv3 first and then Conv2, and each will be refined by the $[3\times3, 256]$ operation. The whole decoding procedure is then similar to the U-Net/SegNet design \cite{ronneberger2015u,badrinarayanan2015segnet}. However, we have not observed significant improvement. Thus, in the end, we adopt the very simple yet effective decoder module: the concatenation of the DeepLabv3 feature map and the channel-reduced Conv2 feature map are refined by two $[3\times3, 256]$ operations. Note that our proposed DeepLabv3+ model has $\emph{output\ stride}=4$. We do not pursue further denser output feature map (\ie, $\emph{output\ stride} < 4$) given the limited GPU resources.

\begin{table}[!t]
  \centering
  \scalebox{1}{
  \begin{tabular}{c | c c c c c}
    \toprule[0.2em]
    Channels & 8 & 16 & 32 & 48 & 64 \\
    \toprule[0.2em]
    mIOU & 77.61\% & 77.92\% & 78.16\% & {\bf 78.21\%} & 77.94\% \\
    \bottomrule[0.1em]
  \end{tabular}
  }
  \caption{PASCAL VOC 2012 \textit{val} set. Effect of decoder $1\times1$ convolution used to reduce the channels of low-level feature map from the encoder module. We fix the other components in the decoder structure as using $[3\times3, 256]$ and Conv2.}
  \label{tab:decoder_bottleneck}
\end{table}

\begin{table}[!t]
  \centering
  \scalebox{1}{
  \begin{tabular}{c c c | c}
    \toprule[0.2em]
    \multicolumn{2}{c}{Features} & $3\times3$ Conv & \multirow{2}{*}{mIOU} \\
    Conv2 & Conv3 & Structure &  \\
    \toprule[0.2em]
    \checkmark &              & $[3\times3, 256]$          & 78.21\% \\
    \checkmark &              & $[3\times3, 256] \times 2$ & {\bf 78.85\%} \\
    \checkmark &              & $[3\times3, 256] \times 3$ & 78.02\% \\
    \checkmark &              & $[3\times3, 128]$          & 77.25\% \\
    \checkmark &              & $[1\times1, 256]$          & 78.07\% \\
    \checkmark & \checkmark   & $[3\times3, 256]$ & 78.61\% \\
    \bottomrule[0.1em]
  \end{tabular}
 }
  \caption{Effect of decoder structure when fixing $[1\times1, 48]$ to reduce the encoder feature channels. We found that it is most effective to use the Conv2 (before striding) feature map and two extra $[3\times3, 256]$ operations. Performance on VOC 2012 \textit{val} set.}
  \label{tab:decoder_structure}
\end{table}

\begin{table*}[!t]
  \centering
  \scalebox{1}{
  \begin{tabular}{c c c c c | c c}
    \toprule[0.2em]
    \multicolumn{2}{c}{Encoder} & \multirow{2}{*}{Decoder} & \multirow{2}{*}{MS} & \multirow{2}{*}{Flip} & \multirow{2}{*}{mIOU} & \multirow{2}{*}{Multiply-Adds} \\
    train OS & eval OS & & & & & \\
    \toprule[0.2em]
    16 & 16 &            &            &            & 77.21\%  & 81.02B \\ 
    16 &  8 &            &            &            & 78.51\%  & 276.18B \\ 
    16 &  8 &            & \checkmark &            & 79.45\%  & 2435.37B \\ 
    16 &  8 &            & \checkmark & \checkmark & 79.77\%  & 4870.59B \\ 
    \midrule
    16 & 16 & \checkmark &            &            & 78.85\%  & 101.28B \\ 
    16 & 16 & \checkmark & \checkmark &            & 80.09\%  & 898.69B \\ 
    16 & 16 & \checkmark & \checkmark & \checkmark & 80.22\%  & 1797.23B \\ 
    16 & 8  & \checkmark &            &            & 79.35\%  & 297.92B \\ 
    16 & 8  & \checkmark & \checkmark &            & 80.43\%  & 2623.61B \\ 
    16 & 8  & \checkmark & \checkmark & \checkmark & 80.57\%  & 5247.07B \\ 
    \midrule
    32 & 32 &            &            &            & 75.43\% & 52.43B \\ 
    32 & 32 & \checkmark &            &            & 77.37\% & 74.20B \\ 
    32 & 16 & \checkmark &            &            & 77.80\% & 101.28B \\ 
    32 & 8  & \checkmark &            &            & 77.92\% & 297.92B \\ 
    \bottomrule[0.1em]
  \end{tabular}
 }
  \caption{Inference strategy on the PASCAL VOC 2012 \textit{val} set using {\it ResNet-101}. {\bf train OS}: The $\emph{output\ stride}$ used during training. {\bf eval OS}: The $\emph{output\ stride}$ used during evaluation. {\bf Decoder}: Employing the proposed decoder structure. {\bf MS}: Multi-scale inputs during evaluation. {\bf Flip}: Adding left-right flipped inputs.}
  \label{tab:resnet_decoder_flops}
\end{table*}

\subsection{ResNet-101 as Network Backbone}
To compare the model variants in terms of both accuracy and speed, we report mIOU and Multiply-Adds in \tabref{tab:resnet_decoder_flops} when using ResNet-101 \cite{he2015deep} as network backbone in the proposed DeepLabv3+ model. Thanks to atrous convolution, we are able to obtain features at different resolutions during training and evaluation using a single model.

{\bf Baseline:} The first row block in \tabref{tab:resnet_decoder_flops} contains the results from \cite{chen2017rethinking} showing that extracting denser feature maps during evaluation (\ie, $\emph{eval\ output\ stride} = 8$) and adopting multi-scale inputs increases performance. Besides, adding left-right flipped inputs doubles the computation complexity with only marginal performance improvement.

{\bf Adding decoder:} The second row block in \tabref{tab:resnet_decoder_flops} contains the results when adopting the proposed decoder structure. The performance is improved from $77.21\%$ to $78.85\%$ or $78.51\%$ to $79.35\%$ when using $\emph{eval\ output\ stride} = 16$ or $8$, respectively, at the cost of about 20B extra computation overhead. The performance is further improved when using multi-scale and left-right flipped inputs.

{\bf Coarser feature maps:} We also experiment with the case when using $\emph{train\ output\ stride} = 32$ (\ie, no atrous convolution at all during training) for fast computation. As shown in the third row block in \tabref{tab:resnet_decoder_flops}, adding the decoder brings about 2\% improvement while only 74.20B Multiply-Adds are required. However, the performance is always about 1\% to 1.5\% below the case in which we employ $\emph{train\ output\ stride} = 16$ and different $\emph{eval\ output\ stride}$ values. We thus prefer using $\emph{output\ stride} = 16$ or $8$ during training or evaluation depending on the complexity budget.

\subsection{Xception as Network Backbone}
We further employ the more powerful Xception \cite{chollet2016xception} as network backbone. Following \cite{dai2017coco}, we make a few more changes, as described in \secref{subsec:xception}.

\begin{table}[!t]
  \centering
  \begin{tabular}{c | c c}
    \toprule[0.2em]
    Model & Top-1 Error & Top-5 Error \\
    \toprule[0.2em]
    Reproduced ResNet-101 & 22.40\% & 6.02\% \\
    Modified Xception & 20.19\% & 5.17\% \\
    \bottomrule[0.1em]
  \end{tabular}
  \caption{{\it Single-model} error rates on ImageNet-1K validation set.}
  \label{tab:xception_imagenet}
\end{table}

{\bf ImageNet pretraining:} The proposed Xception network is pretrained on ImageNet-1k dataset \cite{ILSVRC15} with similar training protocol in \cite{chollet2016xception}. Specifically, we adopt Nesterov momentum optimizer with momentum = 0.9, initial learning rate = 0.05, rate decay = 0.94 every 2 epochs, and weight decay $4e-5$. We use asynchronous training with 50 GPUs and each GPU has batch size 32 with image size $299\times299$. We did not tune the hyper-parameters very hard as the goal is to pretrain the model on ImageNet for semantic segmentation. We report the {\it single-model} error rates on the validation set in \tabref{tab:xception_imagenet} along with the baseline reproduced ResNet-101 \cite{he2015deep} under the same training protocol. We have observed 0.75\% and 0.29\% performance degradation for Top1 and Top5 accuracy when not adding the extra batch normalization and ReLU after each $3\times3$ depthwise convolution in the modified Xception. 

The results of using the proposed Xception as network backbone for semantic segmentation are reported in \tabref{tab:xception_decoder_flops}.

{\bf Baseline:} We first report the results without using the proposed decoder in the first row block in \tabref{tab:xception_decoder_flops}, which shows that employing Xception as network backbone improves the performance by about 2\% when $\emph{train\ output\ stride} = \emph{eval\ output\ stride} = 16$ over the case where ResNet-101 is used. Further improvement can also be obtained by using $\emph{eval\ output\ stride} = 8$, multi-scale inputs during inference and adding left-right flipped inputs. Note that we do not employ the multi-grid method \cite{wang2017understanding,dai2017deformable,chen2017rethinking}, which we found does not improve the performance.

{\bf Adding decoder:} As shown in the second row block in \tabref{tab:xception_decoder_flops}, adding decoder brings about 0.8\% improvement when using $\emph{eval\ output\ stride}=16$ for all the different inference strategies. The improvement becomes less when using $\emph{eval\ output\ stride} = 8$.

{\bf Using depthwise separable convolution:} Motivated by the efficient computation of depthwise separable convolution, we further adopt it in the ASPP and the decoder modules. As shown in the third row block in \tabref{tab:xception_decoder_flops}, the computation complexity in terms of Multiply-Adds is significantly reduced by 33\% to 41\%, while similar mIOU performance is obtained.

{\bf Pretraining on COCO:} For comparison with other state-of-art models, we further pretrain our proposed DeepLabv3+ model on MS-COCO dataset \cite{lin2014microsoft}, which yields about extra 2\% improvement for all different inference strategies.

{\bf Pretraining on JFT:} Similar to \cite{chen2017rethinking}, we also employ the proposed Xception model that has been pretrained on both ImageNet-1k \cite{ILSVRC15} and JFT-300M dataset \cite{hinton2015distilling,chollet2016xception,sun2017revisiting}, which brings extra 0.8\% to 1\% improvement.

{\bf Test set results:} Since the computation complexity is not considered in the benchmark evaluation, we thus opt for the best performance model and train it with $\emph{output\ stride}=8$ and frozen batch normalization parameters. In the end, our `DeepLabv3+' achieves the performance of 87.8\% and 89.0\% without and with JFT dataset pretraining. 

\begin{figure}
  \centering
  \scalebox{1}{
  \begin{tabular}{c c c c c c}
    \includegraphics[width=0.15\linewidth]{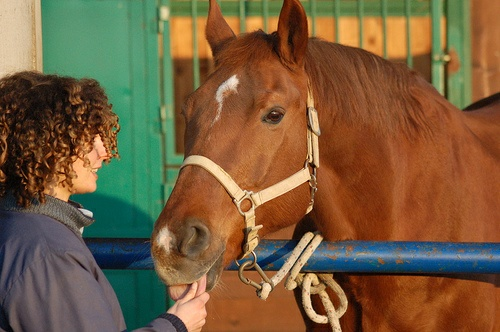} &
    \includegraphics[width=0.15\linewidth]{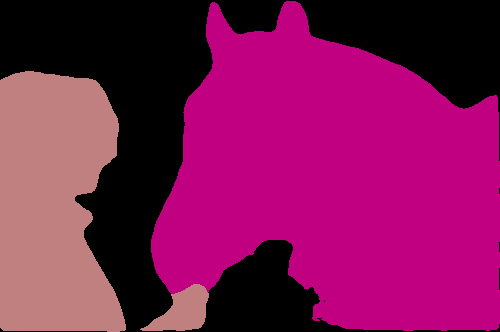} &

    \includegraphics[width=0.15\linewidth]{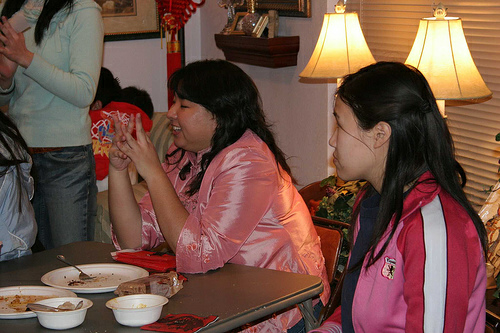} &
    \includegraphics[width=0.15\linewidth]{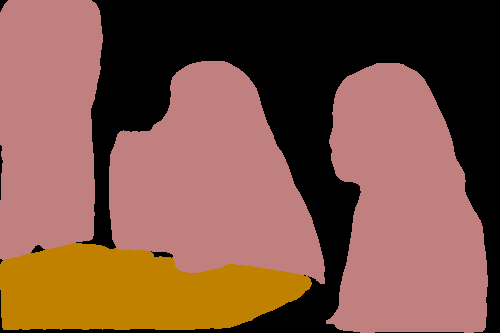} &

    \includegraphics[width=0.15\linewidth]{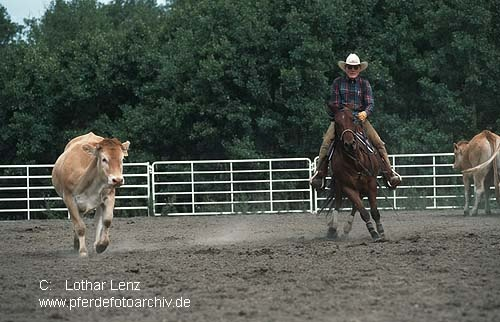} &
    \includegraphics[width=0.15\linewidth]{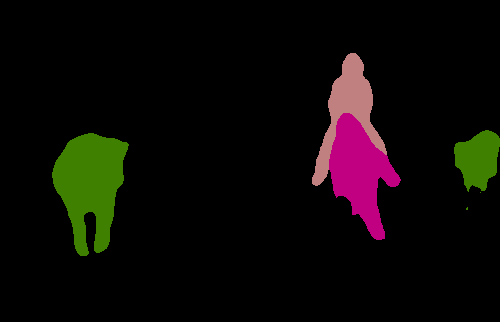} \\

    \includegraphics[width=0.15\linewidth]{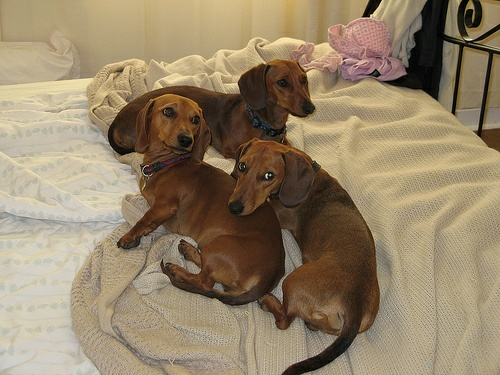} &
    \includegraphics[width=0.15\linewidth]{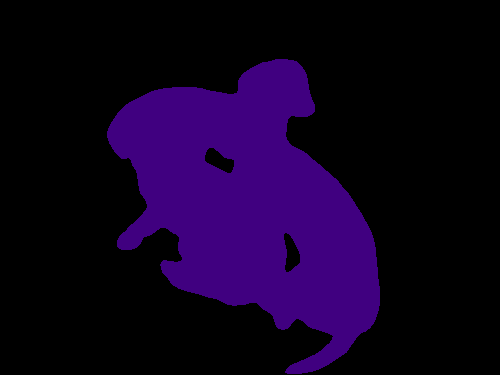} &

    \includegraphics[width=0.15\linewidth]{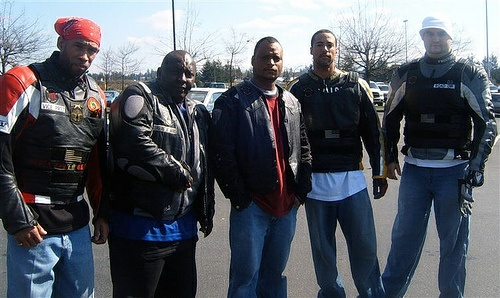} &
    \includegraphics[width=0.15\linewidth]{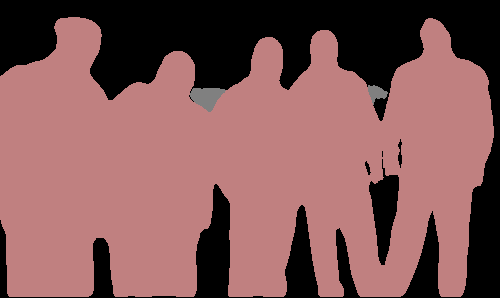} &

    \includegraphics[width=0.15\linewidth]{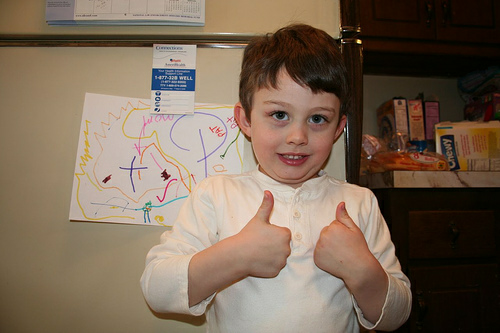} &
    \includegraphics[width=0.15\linewidth]{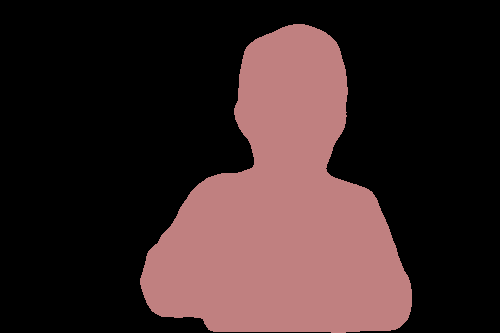} \\

    \toprule[0.1em]
    
    \includegraphics[width=0.15\linewidth]{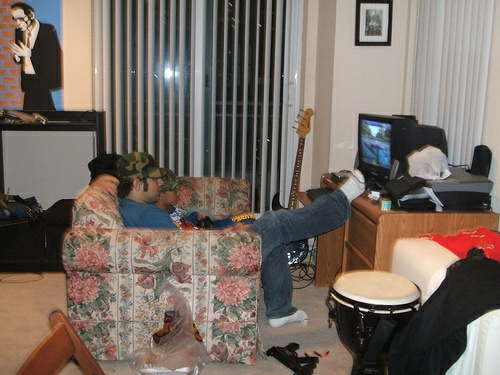} &
    \includegraphics[width=0.15\linewidth]{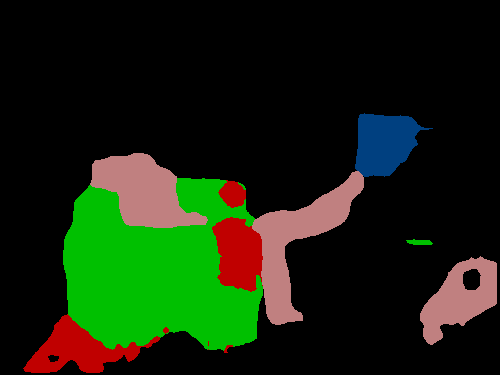} &

    \includegraphics[width=0.15\linewidth]{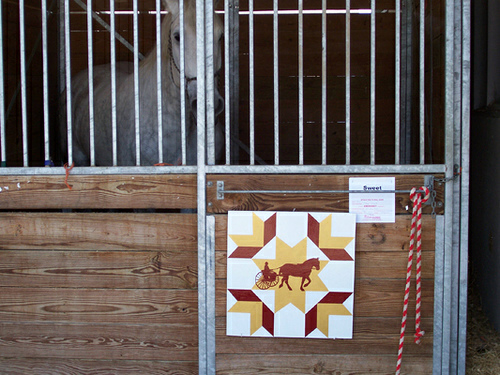} &
    \includegraphics[width=0.15\linewidth]{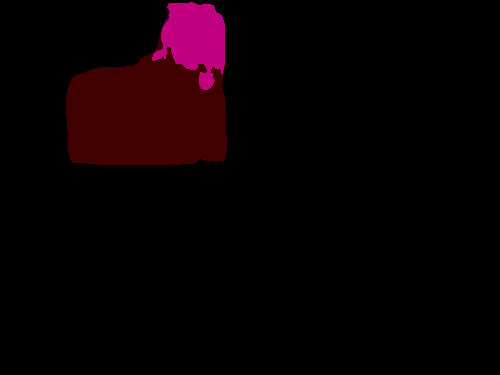} &

    \includegraphics[width=0.15\linewidth]{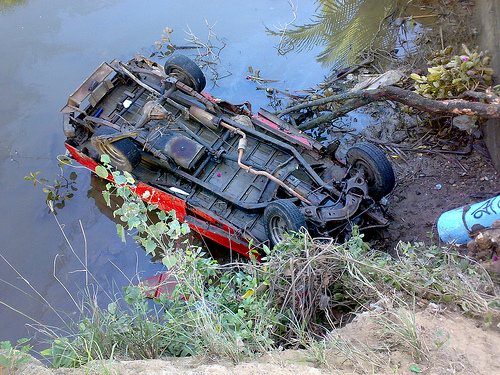} &
    \includegraphics[width=0.15\linewidth]{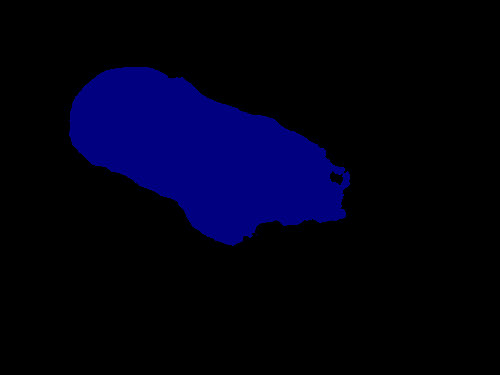} \\
  \end{tabular}
  }
  \caption{Visualization results on \textit{val} set. The last row shows a failure mode.}
  \label{fig:vis_results}
\end{figure}

{\bf Qualitative results:} We provide visual results of our best model in \figref{fig:vis_results}. As shown in the figure, our model is able to segment objects very well without any post-processing.

\textbf{Failure mode:} As shown in the last row of \figref{fig:vis_results}, our model has difficulty in segmenting (a) sofa \vs chair, (b) heavily occluded objects, and (c) objects with rare view.

\begin{table*}[!t]
  \centering
  \scalebox{1}{
  \begin{tabular}{c c c c c c c c | c c}
    \toprule[0.2em]
    \multicolumn{2}{c}{Encoder} & \multirow{2}{*}{Decoder} & \multirow{2}{*}{MS} & \multirow{2}{*}{Flip} & \multirow{2}{*}{SC} & \multirow{2}{*}{COCO} & \multirow{2}{*}{JFT} & \multirow{2}{*}{mIOU} & \multirow{2}{*}{Multiply-Adds} \\
    train OS & eval OS & &            &            &            &            &            &         &         \\
    \toprule[0.2em]
    16 & 16 &            &            &            &            &            &            & 79.17\% & 68.00B \\ 
    16 & 16 &            & \checkmark &            &            &            &            & 80.57\% & 601.74B \\ 
    16 & 16 &            & \checkmark & \checkmark &            &            &            & 80.79\% & 1203.34B \\ 
    16 &  8 &            &            &            &            &            &            & 79.64\% & 240.85B \\ 
    16 &  8 &            & \checkmark &            &            &            &            & 81.15\% & 2149.91B \\ 
    16 &  8 &            & \checkmark & \checkmark &            &            &            & 81.34\% & 4299.68B \\ 
    \midrule
    16 & 16 & \checkmark &            &            &            &            &            & 79.93\% & 89.76B \\ 
    16 & 16 & \checkmark & \checkmark &            &            &            &            & 81.38\% & 790.12B \\ 
    16 & 16 & \checkmark & \checkmark & \checkmark &            &            &            & 81.44\% & 1580.10B \\ 
    16 & 8  & \checkmark &            &            &            &            &            & 80.22\% & 262.59B \\ 
    16 & 8  & \checkmark & \checkmark &            &            &            &            & 81.60\% & 2338.15B \\ 
    16 & 8  & \checkmark & \checkmark & \checkmark &            &            &            & 81.63\% & 4676.16B \\ 
    \midrule
    \midrule
    16 & 16 & \checkmark &            &            & \checkmark &            &            & 79.79\% & 54.17B \\ 
    16 & 16 & \checkmark & \checkmark & \checkmark & \checkmark &            &            & 81.21\% & 928.81B \\ 
    16 & 8  & \checkmark &            &            & \checkmark &            &            & 80.02\% & 177.10B \\ 
    16 & 8  & \checkmark & \checkmark & \checkmark & \checkmark &            &            & 81.39\% & 3055.35B \\ 
    \midrule
    16 & 16 & \checkmark &            &            & \checkmark & \checkmark &            & 82.20\% & 54.17B \\ 
    16 & 16 & \checkmark & \checkmark & \checkmark & \checkmark & \checkmark &            & 83.34\% & 928.81B \\ 
    16 & 8  & \checkmark &            &            & \checkmark & \checkmark &            & 82.45\% & 177.10B \\ 
    16 & 8  & \checkmark & \checkmark & \checkmark & \checkmark & \checkmark &            & 83.58\% & 3055.35B \\ 
    \midrule
    16 & 16 & \checkmark &            &            & \checkmark & \checkmark & \checkmark & 83.03\% & 54.17B \\ 
    16 & 16 & \checkmark & \checkmark & \checkmark & \checkmark & \checkmark & \checkmark & 84.22\% & 928.81B \\ 
    16 & 8  & \checkmark &            &            & \checkmark & \checkmark & \checkmark & 83.39\% & 177.10B \\ 
    16 & 8  & \checkmark & \checkmark & \checkmark & \checkmark & \checkmark & \checkmark & 84.56\% & 3055.35B \\ 
    \bottomrule[0.1em]
  \end{tabular}
  }
  \caption{Inference strategy on the PASCAL VOC 2012 \textit{val} set when using modified {\it Xception}. {\bf train OS}: The $\emph{output\ stride}$ used during training. {\bf eval OS}: The $\emph{output\ stride}$ used during evaluation. {\bf Decoder}: Employing the proposed decoder structure. {\bf MS}: Multi-scale inputs during evaluation. {\bf Flip}: Adding left-right flipped inputs. {\bf SC}: Adopting depthwise separable convolution for both ASPP and decoder modules. {\bf COCO}: Models pretrained on MS-COCO. {\bf JFT}: Models pretrained on JFT.}
  \label{tab:xception_decoder_flops}
\end{table*}

\begin{table}[!t]
  \centering
  \scalebox{0.8}{
  \addtolength{\tabcolsep}{2.5pt}
  \begin{tabular}{l | c}
    \toprule[0.2 em]
    {\bf Method} & {\bf mIOU} \\
    \toprule[0.2 em]
    Deep Layer Cascade (LC) \cite{li2017not} & 82.7 \\
    TuSimple \cite{wang2017understanding} & 83.1 \\
    Large\_Kernel\_Matters \cite{peng2017large} & 83.6 \\
    Multipath-RefineNet \cite{lin2016refinenet} & 84.2 \\
    ResNet-38\_MS\_COCO \cite{wu2016wider} & 84.9 \\
    PSPNet \cite{zhao2017pyramid} & 85.4 \\
    IDW-CNN \cite{wanglearning}  & 86.3 \\
    CASIA\_IVA\_SDN \cite{fu2017stacked} & 86.6 \\
    DIS \cite{luo2017deep} & 86.8 \\
    \midrule
    DeepLabv3 \cite{chen2017rethinking} & 85.7 \\
    DeepLabv3-JFT \cite{chen2017rethinking} & 86.9 \\
    \midrule
    \href{http://host.robots.ox.ac.uk:8080/anonymous/NU9OS6.html}{DeepLabv3+ (Xception)} & 87.8 \\
    \href{http://host.robots.ox.ac.uk:8080/anonymous/AF0NVP.html}{DeepLabv3+ (Xception-JFT)} & 89.0 \\
    \bottomrule[0.1 em]
  \end{tabular}
  }
  \caption{PASCAL VOC 2012 {\it test} set results with top-performing models.}
  \label{tab:xception_testset}
\end{table}

\subsection{Improvement along Object Boundaries}
\begin{figure}[t!]
  \begin{tabular}{c c}
    \scalebox{0.45}{
      \begin{tabular}{c}
        \includegraphics[width=0.98\linewidth]{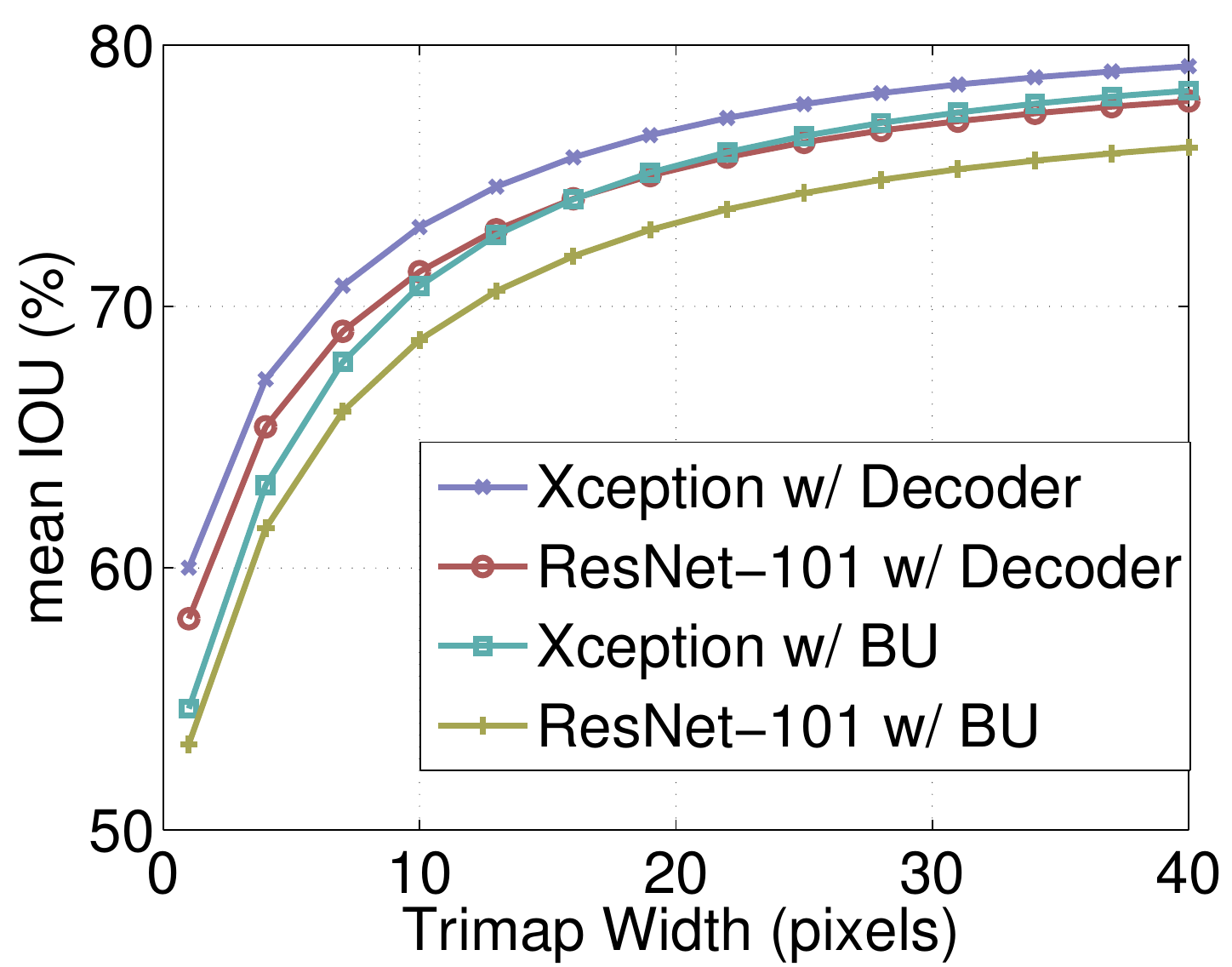} \\
      \end{tabular}
    }
    &
    \scalebox{0.58}{
      \begin{tabular}{c c c}
        \includegraphics[width=0.29\linewidth]{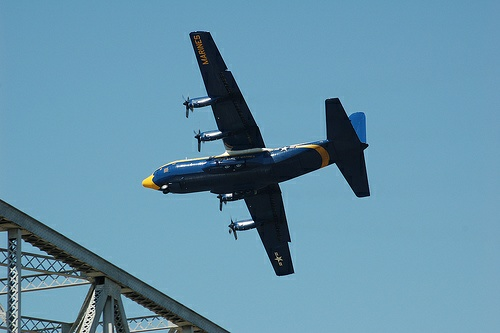} &
        \includegraphics[width=0.29\linewidth]{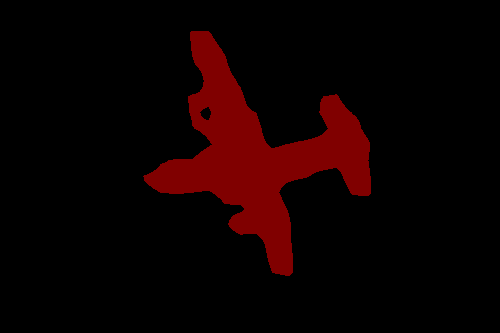} &
        \includegraphics[width=0.29\linewidth]{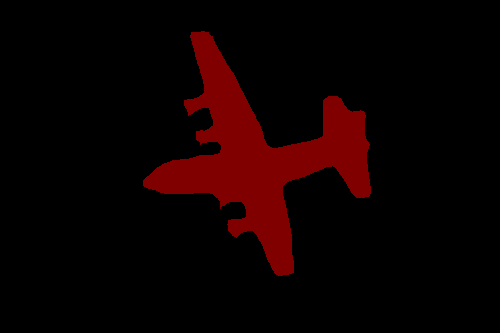} \\
        \includegraphics[width=0.29\linewidth]{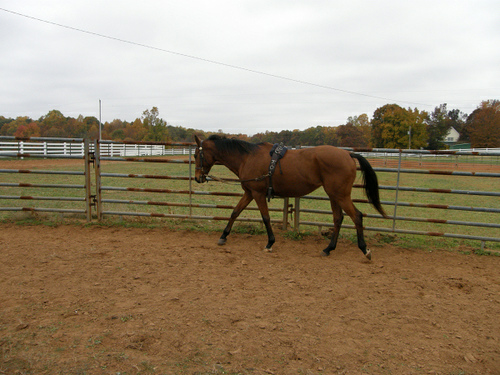} &
        \includegraphics[width=0.29\linewidth]{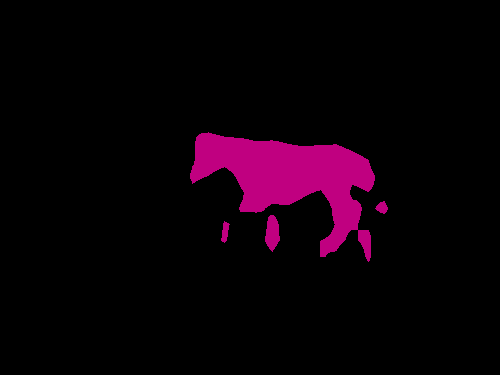} &
        \includegraphics[width=0.29\linewidth]{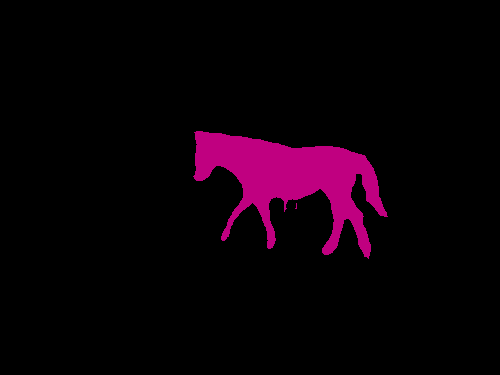} \\
                    {Image} &
                    {w/ BU} &
                    {w/ Decoder} \\
      \end{tabular}
    }
    \\
    (a) mIOU \vs Trimap width & (b) Decoder effect \\
  \end{tabular}
  \caption{(a) mIOU as a function of trimap band width around the object boundaries when employing $\emph{train\ output\ stride} = \emph{eval\ output\ stride} = 16$. {\bf BU}: Bilinear upsampling. (b) Qualitative effect of employing the proposed decoder module compared with the naive bilinear upsampling (denoted as {\bf BU}). In the examples, we adopt Xception as feature extractor and $\emph{train\ output\ stride} = \emph{eval\ output\ stride} = 16$.}
  \label{fig:trimap_and_decoder}
\end{figure}

In this subsection, we evaluate the segmentation accuracy with the trimap experiment \cite{kohli2009robust,krahenbuhl2011efficient,chen2017deeplab} to quantify the accuracy of the proposed decoder module near object boundaries. Specifically, we apply the morphological dilation on `void' label annotations on \textit{val} set, which typically occurs around object boundaries. We then compute the mean IOU for those pixels that are within the dilated band (called trimap) of `void' labels. As shown in \figref{fig:trimap_and_decoder}~(a), employing the proposed decoder for both ResNet-101 \cite{he2015deep} and Xception \cite{chollet2016xception} network backbones improves the performance compared to the naive bilinear upsampling. The improvement is more significant when the dilated band is narrow. We have observed 4.8\% and 5.4\% mIOU improvement for ResNet-101 and Xception respectively at the smallest trimap width as shown in the figure. We also visualize the effect of employing the proposed decoder in \figref{fig:trimap_and_decoder}~(b).

\subsection{Experimental Results on Cityscapes}
In this section, we experiment DeepLabv3+ on the Cityscapes dataset \cite{Cordts2016Cityscapes}, a large-scale dataset containing high quality pixel-level annotations of 5000 images (2975, 500, and 1525 for the training, validation, and test sets respectively) and about 20000 coarsely annotated images. 

As shown in \tabref{tab:cityscapes_val_test}~(a), employing the proposed Xception model as network backbone (denoted as X-65) on top of DeepLabv3 \cite{chen2017rethinking}, which includes the ASPP module and image-level features \cite{liu2015parsenet}, attains the performance of 77.33\% on the validation set. Adding the proposed decoder module significantly improves the performance to 78.79\% (1.46\% improvement). We notice that removing the augmented image-level feature improves the performance to 79.14\%, showing that in DeepLab model, the image-level features are more effective on the PASCAL VOC 2012 dataset. We also discover that on the Cityscapes dataset, it is effective to increase more layers in the entry flow in the Xception \cite{chollet2016xception}, the same as what \cite{dai2017coco} did for the object detection task. The resulting model building on top of the deeper network backbone (denoted as X-71 in the table), attains the best performance of 79.55\% on the validation set.

After finding the best model variant on \textit{val} set, we then further fine-tune the model on the coarse annotations in order to compete with other state-of-art models. As shown in \tabref{tab:cityscapes_val_test}~(b), our proposed DeepLabv3+ attains a performance of 82.1\% on the test set, setting a new state-of-art performance on Cityscapes.

\begin{table}[t!]
  \begin{tabular}{c c}
    \scalebox{1}{
      \begin{tabular}{c c c c | c}
        \toprule[0.2em]
        Backbone & Decoder & ASPP & Image-Level & mIOU \\
        \toprule[0.2em]
        X-65 &              & \checkmark & \checkmark & 77.33 \\
        X-65 & \checkmark   & \checkmark & \checkmark & 78.79 \\
        X-65 & \checkmark   & \checkmark &            & 79.14 \\
        X-71 & \checkmark   & \checkmark &            & 79.55 \\
        \bottomrule[0.1em]
      \end{tabular}
    }
  &
  {
    \addtolength{\tabcolsep}{2.5pt}
    \begin{tabular}{l c | c}
      \toprule[0.2 em]
              {\bf Method} & {\bf Coarse} & {\bf mIOU} \\
              \toprule[0.2 em]
              ResNet-38 \cite{wu2016wider} & \checkmark & 80.6 \\
              PSPNet \cite{zhao2017pyramid} & \checkmark & 81.2 \\
              Mapillary \cite{bulo2017place} & \checkmark & 82.0 \\
              \midrule
              DeepLabv3 & \checkmark & 81.3 \\
              \midrule
              DeepLabv3+ & \checkmark & 82.1 \\
              \bottomrule[0.1 em]
    \end{tabular}
  } \\
  (a) \textit{val} set results & (b) \textit{test} set results \\
  \end{tabular}
  \caption{(a) DeepLabv3+ on the Cityscapes \textit{val} set when trained with \textit{train\_fine} set. (b) DeepLabv3+ on Cityscapes {\it test} set. {\bf Coarse}: Use \textit{train\_extra} set (coarse annotations) as well. Only a few top models are listed in this table.}
  \label{tab:cityscapes_val_test}
\end{table}

\section{Conclusion}
Our proposed model ``DeepLabv3+'' employs the encoder-decoder structure where DeepLabv3 is used to encode the rich contextual information and a simple yet effective decoder module is adopted to recover the object boundaries. One could also apply the atrous convolution to extract the encoder features at an arbitrary resolution, depending on the available computation resources. We also explore the Xception model and atrous separable convolution to make the proposed model faster and stronger. Finally, our experimental results show that the proposed model sets a new state-of-the-art performance on PASCAL VOC 2012 and Cityscapes datasets.

\textbf{Acknowledgments}
We would like to acknowledge the valuable discussions with Haozhi Qi and Jifeng Dai about Aligned Xception, the feedback from Chen Sun, and the support from Google Mobile Vision team.

\clearpage

\bibliographystyle{splncs}
\bibliography{egbib}

\end{document}